\documentclass[conference]{IEEEtran}
\IEEEoverridecommandlockouts
\usepackage{cite}
\usepackage{amsmath,amssymb,amsfonts}
\usepackage{algorithmic}
\usepackage{booktabs}
\usepackage{graphicx}
\usepackage{caption}
\usepackage{subcaption}
\usepackage{textcomp}
\usepackage{multirow}
\usepackage{makecell}
\usepackage{amsmath, amssymb}
\usepackage{array}
\usepackage{xcolor}
\def\BibTeX{{\rm B\kern-.05em{\sc i\kern-.025em b}\kern-.08em
    T\kern-.1667em\lower.7ex\hbox{E}\kern-.125emX}}
\begin{document}

\title{An Adaptive Framework for Multi-View Clustering Leveraging Conditional Entropy Optimization 
% \\ This work is supported in part by the Science and Technology Development Fund, Macau SAR, under Grant 0087/2020/A2 and Grant 0141/2023/RIA2.
}
% \author{
%     Lijian Li\textsuperscript{1}, Yuanpeng He\textsuperscript{2, 3}, Chi-Man Pun\textsuperscript{1} \\
%     \textsuperscript{1}Department of Computer and Information Science, University of Macau, Macau, China \\
%     \textsuperscript{2}Key Laboratory of High Confidence Software Technologies (Peking University),\\ Ministry of Education, Beijing, 100871, China \\
%     \textsuperscript{3}School of Computer Science, Peking University, Beijing, 100871, China \\
%     mc35305@umac.mo, cmpun@umac.mo
% }

\author{\IEEEauthorblockN{Lijian Li}
% \and
% \and
% \and
}

\maketitle

\begin{abstract}
Multi-view clustering (MVC) has emerged as a powerful technique for extracting valuable insights from data characterized by multiple perspectives or modalities. Despite significant advancements, existing MVC methods struggle with effectively quantifying the consistency and complementarity among views, and are particularly susceptible to the adverse effects of noisy views, known as the Noisy-View Drawback (NVD). To address these challenges, we propose CE-MVC, a novel framework that integrates an adaptive weighting algorithm with a parameter-decoupled deep model. Leveraging the concept of conditional entropy and normalized mutual information, CE-MVC quantitatively assesses and weights the informative contribution of each view, facilitating the construction of robust unified representations. The parameter-decoupled design enables independent processing of each view, effectively mitigating the influence of noise and enhancing overall clustering performance. Extensive experiments demonstrate that CE-MVC outperforms existing approaches, offering a more resilient and accurate solution for multi-view clustering tasks.
\end{abstract}


\section{Introduction}
In recent times, multi-view data has become increasingly prevalent, with each instance being described from multiple perspectives or modalities \cite{he2024generalized}. However, the lack of reliable labeled data makes it challenging to extract useful information from multi-view data. To address this issue, researchers have proposed an effective self-supervised clustering technique known as multi-view clustering (MVC), which is used to identify pattern structures in various unlabeled data, particularly in fields such as medical image analysis and drug discovery \cite{DBLP:journals/tcyb/XiaZGSHG22, DBLP:journals/tsmc/WenZFZXZL23}. Numerous methods \cite{DBLP:conf/cvpr/Wen0X0HF023, DBLP:conf/iccv/GaoNLH15, DBLP:journals/tnn/WangWLG18} based on graph learning, subspace learning, and matrix factorization have been introduced and achieved some success with multi-view data. Additionally, deep learning-based MVC methods \cite{DBLP:conf/cikm/NigamG00, DBLP:journals/corr/abs-1807-03748, DBLP:journals/corr/abs-2206-07579, DBLP:conf/cvpr/Tang022, DBLP:conf/icml/TangL22}, utilizing self-supervised techniques such as adversarial learning and self-training, have also garnered significant attention. 

The integration of evidence-based learning approaches \cite{he2022mmget,  he2023tdqmf} into MVC has proven useful in enhancing the accuracy and robustness of clustering. By leveraging evidence from different modalities or views \cite{he2021conflicting, xu2023spatio, he2022new}, the model can better understand the underlying relationships between data points, which is particularly important for classification tasks \cite{li2022nndf, he2021new} where precise decision boundaries are essential. In particular, evidence aggregation strategies, such as those based on consensus or majority voting, have been shown to improve the final cluster quality by reducing the impact of noisy or ambiguous data from individual views.

Furthermore, the application of visual models in MVC has seen notable advancements \cite{DBLP:conf/icml/Liu0LWZTT0Z21, DBLP:conf/ijcai/WangLZTLHXY19}. Convolutional neural networks (CNNs) \cite{he2024residual} and transformer-based architectures, which are well-known for their strength in visual perception tasks, have been incorporated into multi-view clustering frameworks to handle image data. These models not only help extract rich visual features from each view but also facilitate the alignment of information across views, contributing to more accurate and consistent clustering results. The synergy between visual models and MVC can particularly benefit domains like medical imaging, where fine-grained classification and robust clustering are critical for tasks such as disease detection or organ segmentation.

The effectiveness of current MVC approaches stems from their ability to leverage both consistency and complementarity \cite{DBLP:journals/pami/ZhangFHCXTX20, DBLP:journals/pami/LiuZLWTYSWG19, DBLP:conf/cvpr/Tang022} inherent in multi-view data, leading to superior performance compared to single-view clustering (SVC) methods. Consistency refers to the shared information across multiple views that aids in identifying the same category. For instance, consistent categorical information across different views can enhance the understanding of category semantics, thereby reducing the impact of irrelevant or non-semantic data. Complementarity, on the other hand, suggests that diverse views offer supplementary information that can mutually correct and enhance each other \cite{DBLP:journals/pami/LiuLYLX23, DBLP:journals/tkde/TangLWLZZ23}. By integrating these different perspectives, it becomes possible to uncover category structures that remain hidden when considering individual views alone. Despite these advantages, a significant challenge remains: the concepts of consistency and complementarity across multiple views are still conceptually abstract and difficult to quantify. Beyond the abstract attributes of multi-view data, the presence of noisy views significantly exacerbates the complexity of clustering tasks. Unlike informative views that contribute complementary insights, thereby facilitating the accurate identification of shared categories, the information derived from noisy views is not only irrelevant but may also mislead existing semantic representations, ultimately resulting in a deterioration of clustering performance, which is entitled as Noisy-View Drawback (NVD). For existing MVC methods \cite{DBLP:conf/ijcai/WangZWLFP16}, NVD undermines their effectiveness because they often rely on shared neural networks to fuse representations across views. When clustering objectives prioritize the noisy view, the shared parameters tend to overfit this view, neglecting valuable information from more informative views. Additionally, enforcing consistent clustering predictions across all views \cite{nie2016parameter, DBLP:conf/cvpr/Tang022, DBLP:journals/tnn/WangWLG18, DBLP:conf/cvpr/JinWDLZ23, DBLP:conf/cvpr/ZhouS20}, including the noisy ones, can degrade the overall representation learning and clustering performance.

In this paper, to address these issues, we propose a quantifiable multi-view clustering framework, CE-MVC, which integrates an adaptive weighting algorithm with a parameter-decoupled deep model. Inspired by entropy \cite{DBLP:journals/eswa/ZhaoNMZ24, he2023ordinal, DBLP:journals/paa/ZhuM20, DBLP:journals/jmihi/WuSL20, DBLP:journals/tcsv/WangCLCC23, he2022ordinal} as a measure of information content, we design a conditional entropy metric to quantify complementary information within each view. This metric, combined with NMI, is used to weigh views during the formation of a unified representation. To address the Noisy-View Drawback (NVD) issue, we propose a parameter-decoupled deep model that processes data representations and generates soft labels for different views, aiming to effectively mitigate the adverse impact of noisy views on the overall clustering performance.
% we proposed a framework utilize deep models with separate parameters to independently learn representations and generate soft labels for each view, with the objective of mitigating the adverse impacts caused by noisy views.
% The effectiveness of current MVC approaches comes from leveraging two key properties: consistency and complementarity in multi-view data. Consistency involves shared information across views that helps in identifying categories, enhancing understanding by reducing the influence of irrelevant data. Complementarity, on the other hand, suggests that different views offer supplementary information, allowing them to mutually improve and reveal hidden category structures. However, these concepts remain abstract and challenging to measure.

\section{Methods}
\subsection{Annotation}
We consider a multi-view dataset denoted as $\{\mathbf{X}^v \in \mathbb{R}^{N\times D_v}\}^V_{v=1}$, comprising $N$ samples observed from $V$ distinct views. For each view $v$, the learned feature representations and corresponding soft labels are represented by $\mathbf{R}^v \in \mathbb{R}^{N\times d_v}$ and $\mathbf{SL}^v \in \mathbb{R}^{N\times K}$, respectively. $D_v, d_v$ indicate the dimensionality of input data $\mathbf{X}^v$ and the latent representation $\mathbf{R}^v$ for the $v$-th view. The parameter $K$ refers to the number of clusters.

\subsection{Preliminary}
The learning paradigm in MVC typically involves learning feature representations $\mathbf{R}$ and soft labels $\mathbf{SL}$, followed by using fusion strategies ($e.g.$, early or late fusion) to leverage valuable information across views. Despite advancements in weighting strategies to account for view diversity, most MVC methods rely on shared network parameters and consistent clustering across views, which can lead to reduced robustness, especially when faced with low-quality or noisy views. The common clustering objective for them is defined as follows:
\begin{equation}
    \min_{\mathbf{\Theta}} \sum_{v=1}^{V} \|\mathbf{T} - \mathcal{F}_{\mathbf{\Theta}}(\mathbf{SL}^v \mid \{\mathbf{R}^v\}_{v=1}^V)\|_F^2,
\end{equation}
where $\{\mathbf{R}^v\}^V_{v=1}$ denotes feature representations of all views and $\mathcal{F}_{\Theta}(\mathbf{Y}^v \mid \{\mathbf{R}^v\}_{v=1}^V)$ indicates that soft labels are derived through the fusion module $\mathcal{F}$ applied to the representations $\{\mathbf{R}^v\}_{v=1}^V$ using the parameter set $\mathbf{\Theta}$ shared across all views. The matrix $\mathbf{T}$ serves as a unified learning target to ensure consistency in the soft labels $\{\mathbf{SL}^v\}_{v=1}^V$ across all views.
\subsection{Asymptotic adaptive weighting optimization based on conditional entropy}
To eliminate the reliance on shared networks and parameters, we propose an asymptotic adaptive weighting strategy based on conditional entropy to effectively leverage the complementary and consistent information in multi-view data. This strategy involves learning a reliable target $\mathbf{T}$ while keeping model parameters fixed, thereby ensuring the condition of parameter decoupling is met. We define a weighting matrix $\mathbf{W}_{(t)}$ to automatically explore the complementary information within each view at $t$-th iteration, quantified by specifically designed conditional entropy and NMI metrics. Subsequently, The weighting matrix is then used to produce well-scaled feature representations $\mathbf{R}_{(t)}$, which are utilized to generate reliable soft labels $\mathbf{SL}_{(t)}$ and learning targets $\mathbf{T}_{(t)}$.

Specifically, in the initial iteration, we assume that the amount of effective information across all views is equivalent. Therefore, we initialize a diagonal weighting matrix with values set to 1, corresponding to the weight of each view. This weighting matrix $\mathbf{W}_{(t)}\in \mathbb{R}^{\sum_v d_v \times \sum_v d_v}$ is then used to generate scaled feature representation $\mathbf{R}_{(t)} \in \mathbb{R}^{N\times \sum_v d_v}$ by adjusting the latent representations of each view accordingly, which is formulated as:
\begin{equation}
    \mathbf{R}_{(0)} = \mathcal{O}(\mathbf{W}_{(0)} \mid \mathbf{R}_{(t)}, \mathbf{R}^1, \mathbf{R}^2, \cdots, \mathbf{R}^V)
\end{equation}
In the $t$-th iteration, We first utilize weight matrix $\mathbf{W}_{(t)}$ generated from $t-1$-th iteration to weight current feature representations of each view. Then, the optimized feature representations $\mathbf{R}_t$ are employed to further refine weight matrix $\mathbf{W}_{(t)}$. The conditional entropy for $\mathbf{R}_t^v$ is calculated. When calculating the entropy of feature representations, we typically use kernel density estimation (KDE) to estimate the entropy. Therefore, the entropy formula based on KDE is defined as follows:
\begin{equation}
\mathcal{E}(\mathbf{R}) = -\int p(\mathbf{R}) \log p(\mathbf{R}) \, dR
\end{equation}
where $p(\mathbf{R})$ denotes the probability density function of feature representation. 
To compute the conditional entropy for the current view $\mathbf{R}_t^v$, we first concatenate its feature representation with that of another view $\mathbf{R}_t^u$ and calculate the entropy of the combined representation, $\mathcal{E}(\mathbf{R}_t^v, \mathbf{R}_t^u)$. We then subtract the entropy of $\mathbf{R}_t^u$ alone, $\mathcal{E}(\mathbf{R}_t^u)$, to obtain the partial conditional entropy for $\mathbf{R}_t^v$ relative to $\mathbf{R}_t^u$. The total conditional entropy for $\mathbf{R}_t^v$, denoted as $\mathcal{E}(\mathbf{R}_t^v \mid \{\mathbf{R}_t^u\}_{u \neq v})$, is computed by summing all these partial conditional entropies:
% We concatenate the feature representation of current view, denoted as $\mathbf{R}_t^v$, with the feature representation of another view, denoted as $\mathbf{R}_t^u$, and then compute the information entropy of the concatenated representation: $\mathcal{E}(\mathbf{R}_t^v, \mathbf{R}_t^u)$, where $\mathcal{E}$ denotes information entropy. Next, we subtract the entropy of the feature representation of the other view alone:$\mathcal{E}(\mathbf{R}_t^u)$. The difference between these two entropies yields the partial conditional entropy for the current view with respect to the other view: $\mathcal{E}(\mathbf{R}_t^v, \mathbf{R}_t^u) - \mathcal{E}(\mathbf{R}_t^u)$
% The total conditional entropy for the current view, denoted as $\mathcal{E}(\mathbf{R}_t^v | \{\mathbf{R}_t^u\}_{u \neq v})$, is the sum of all these partial conditional entropies:
\begin{equation}
    \mathcal{E}(\mathbf{R}_t^v | \{\mathbf{R}_t^u\}_{u \neq v}) = \sum_{u \neq v} \left[ \mathcal{E}(\mathbf{R}_t^v, \mathbf{R}_t^u) - \mathcal{E}(\mathbf{R}_t^u) \right]
\end{equation}
If the current view contains complementary information that is useful for the other views, its conditional entropy will be relatively low. Conversely, if the current view is a noisy one, its conditional entropy will be the highest, as it lacks any complementary information. Thus, conditional entropy effectively quantifies the extent of complementary information shared among different views.

Except for the derivation of conditional entropy, the scale feature representation $\mathbf{R}_t$ is also utilized to generate reliable soft labels $\mathbf{SL}_{(t)} \in \mathbb{R}^{N \times K}$, which encapsulates the shared cluster structure of $\mathbf{R}_t$. Previous approaches predominantly utilize NMI to assess the similarity between clustering outcomes from individual views and the unified clustering result, subsequently applying it as a weighting factor. However, NMI primarily captures the consistency between clustering results, thereby failing to capture the complementary information across different views. Relying exclusively on NMI for weighting might lead to the underestimation of views that, despite having lower NMI scores, contribute significant complementary information, potentially overlooking critical data that can enhance clustering performance. To address this limitation, we propose integrating conditional entropy with NMI to generate a weight matrix $\mathbf{W}_{(t+1)} = \mathcal{K}(\mathbf{W}_{(t+1)} \mid \mathbf{SL}_{(t)}, \mathbf{SL}^1, \mathbf{SL}^2, \cdots, \mathbf{SL}^V)$ that simultaneously accounts for both consistency and complementarity of views, which is formulated as follows:
\begin{equation}
w^v_{(t+1)} = \frac{\exp(\frac{2M(\mathbf{SL}^v;\mathbf{SL}_{(t)})}{E(\mathbf{SL}^v) + E(\mathbf{SL}_{(t)})}) -1}{Norm(\mathcal{E}(\mathbf{R}_t^v | \{\mathbf{R}_t^u\}_{u \neq v}))} \in \mathbf{W}_{(t+1)}
\end{equation}
where $M, E$ denotes mutual information and entropy, and $Norm$ represents normalization. The rationale for using conditional entropy in the denominator lies in the observation that views offering significant complementary information are characterized by lower conditional entropy, while those with less complementary information exhibit higher conditional entropy. Consequently, when a view demonstrates limited consistency but substantial complementarity, the division by a relatively low conditional entropy serves to amplify its weight, thereby effectively balancing and integrating both consistency and complementarity in the weighting process.

\subsection{Parameter-decoupled models}
Previous asymptotic adaptive weighting optimization is designed to learn a reliable learning objective $T$ while keeping model parameters fixed. Subsequently, within the parameter-decoupled model, we aim to train different parameters $\{\mathbf{\Theta}^v\}^V_{v=1}$ for individual views by optimizing the clustering objective, which is defined as follows:
\begin{equation}
    \mathcal{L}_c= \min_{\mathbf{\Theta}^v} \|\mathbf{T}^v - \mathcal{F}_{\mathbf{\Theta}^v}(\mathbf{SL}^v|\mathbf{R}^v)\|_F^2.
\end{equation}
Building on some existing deep MVC methods\cite{DBLP:conf/icml/TangL22, DBLP:journals/tip/WangDTGF21, DBLP:conf/mm/0001ZZWFXZ20, DBLP:journals/tkde/XieLQLZMWT21, DBLP:journals/tkde/XuRTYPYPYH23}, we employ deep autoencoders—a widely used self-supervised representation learning technique—to extract new representations from multi-view data. $\mathcal{H}_{\mathbf{\Omega}^v}$ and $\mathcal{D}_{\mathbf{\Delta}^v}$ represent the encoder and decoder for the $v$-th view, respectively. Due to the unshared network parameters $\mathbf{\Omega}^v$ and $\mathbf{\Delta}^v$ for each view, the reconstruction $\mathbf{\Tilde{X}}^v = \mathcal{D}_{\mathbf{\Delta}^v}(\mathcal{H}_{\mathbf{\Omega}^v}(\mathbf{X}^v))$ relies solely on its representation $\mathbf{R}^v = \mathcal{H}_{\mathbf{\Omega}^v}(\mathbf{X}^v)$. Therefore, the corresponding representation learning objective $\mathcal{L}_r^v$ is defined as follows:
\begin{equation}
    \mathcal{L}_r^v = \min_{\{\mathbf{\Omega}^v, \mathbf{\Delta}^v\}} \|\mathbf{X}^v - \mathcal{D}_{\mathbf{\Delta}^v}(\mathcal{H}_{\mathbf{\Omega}^v}(\mathbf{X}^v))\|_F^2.
\end{equation}
And the loss function for training the parameter-decoupled model of each view is composed of two parts:
\begin{equation}
\mathcal{L}^v = \mathcal{L}_r^v + \lambda \mathcal{L}_c^v.
\end{equation}
where The parameter $\lambda$ balances the trade-off between the reconstruction loss $\mathcal{L}_r^v$ and the clustering loss $\mathcal{L}_c^v$. In this context, different views can't interfere with each other during the training of their respective network parameters, which can be formulated as:$ \{\mathbf{\Theta}^i, \mathbf{\Omega}^i, \mathbf{\Delta}^i\} \cap \{\mathbf{\Theta}^j, \mathbf{\Omega}^j, \mathbf{\Delta}^j\} = \emptyset, \forall i \neq j, \, i, j \in \{1, 2, \dots, V\}$. In this process, parameter-decoupled models further refine the representations and soft labels $\{\mathbf{R}^v, \mathbf{SL}^v \}^V_{v=1}$, which are then used to improve the learning target $\mathbf{T}$. Finally, the clustering results for all multi-view data are produced by $\mathbf{SL}_{(t)}$.

\begin{table*}[ht] \small
\centering
\caption{Comparison of different methods on normal datasets and their noisy-contaminated datasets.}
\label{tab:comparison}
\setlength{\tabcolsep}{3mm}
\begin{tabular}{lcccccccc}
\hline
\multirow{2}{*}{Method} & \multicolumn{2}{c}{NoisyDIGIT} & \multicolumn{2}{c}{DIGIT} & \multicolumn{2}{c}{COIL} & \multicolumn{2}{c}{NoisyCOIL} \\
\cmidrule(lr){2-3} \cmidrule(lr){4-5} \cmidrule(lr){6-7} \cmidrule(lr){8-9}
 & ACC & NMI & ACC & NMI & ACC & NMI & ACC & NMI \\
\hline
DEC-BestV \cite{DBLP:conf/icml/XieGF16} & \multicolumn{1}{l}{80.9} & \multicolumn{1}{l}{78.9} & \multicolumn{1}{l}{80.9} & \multicolumn{1}{l}{78.9} & \multicolumn{1}{l}{76.6} & \multicolumn{1}{l}{81.5} & \multicolumn{1}{l}{76.6} & \multicolumn{1}{l}{81.5} \\
DEC-WorstV \cite{DBLP:conf/icml/XieGF16} & \multicolumn{1}{l}{12.4$_{\tiny\textcolor{green}{-68.5}}$} & \multicolumn{1}{l}{0.4$_{\tiny\textcolor{green}{-78.5}}$} & \multicolumn{1}{l}{54.8$_{\tiny\textcolor{green}{-26.1}}$} & \multicolumn{1}{l}{64.1$_{\tiny\textcolor{green}{-14.8}}$} & \multicolumn{1}{l}{73.5$_{\tiny\textcolor{green}{-3.1}}$} & \multicolumn{1}{l}{77.4$_{\tiny\textcolor{green}{-4.1}}$} & \multicolumn{1}{l}{16.4$_{\tiny\textcolor{green}{-60.2}}$} & \multicolumn{1}{l}{2.8$_{\tiny\textcolor{green}{-78.7}}$} \\
DMJC \cite{DBLP:journals/tkde/XieLQLZMWT21} & \multicolumn{1}{l}{80.7$_{\tiny\textcolor{green}{-0.2}}$} & \multicolumn{1}{l}{82.8$_{\tiny\textcolor{red}{+3.9}}$} & \multicolumn{1}{l}{97.6$_{\tiny\textcolor{red}{+16.7}}$} & \multicolumn{1}{l}{96.2$_{\tiny\textcolor{red}{+17.3}}$} & \multicolumn{1}{l}{91.3$_{\tiny\textcolor{red}{+14.7}}$} & \multicolumn{1}{l}{93.8$_{\tiny\textcolor{red}{+12.3}}$} & \multicolumn{1}{l}{85.5$_{\tiny\textcolor{red}{+9.0}}$} & \multicolumn{1}{l}{92.1$_{\tiny\textcolor{red}{+10.6}}$} \\
DIMC-net \cite{DBLP:conf/mm/0001ZZWFXZ20} & \multicolumn{1}{l}{71.6$_{\tiny\textcolor{green}{-9.3}}$} & \multicolumn{1}{l}{76.5$_{\tiny\textcolor{green}{-2.4}}$} & \multicolumn{1}{l}{90.4$_{\tiny\textcolor{red}{+9.5}}$} & \multicolumn{1}{l}{87.3$_{\tiny\textcolor{red}{+8.4}}$} & \multicolumn{1}{l}{98.5$_{\tiny\textcolor{red}{+21.9}}$} & \multicolumn{1}{l}{97.5$_{\tiny\textcolor{red}{+16.0}}$} & \multicolumn{1}{l}{87.5$_{\tiny\textcolor{red}{+10.9}}$} & \multicolumn{1}{l}{91.8$_{\tiny\textcolor{red}{+10.3}}$} \\
GP-MVC \cite{DBLP:journals/tip/WangDTGF21} & \multicolumn{1}{l}{49.1$_{\tiny\textcolor{green}{-31.8}}$} & \multicolumn{1}{l}{63.5$_{\tiny\textcolor{green}{-15.4}}$} & \multicolumn{1}{l}{58.6$_{\tiny\textcolor{green}{-22.3}}$} & \multicolumn{1}{l}{69.8$_{\tiny\textcolor{green}{-9.1}}$} & \multicolumn{1}{l}{86.1$_{\tiny\textcolor{red}{+9.5}}$} & \multicolumn{1}{l}{77.5$_{\tiny\textcolor{green}{-4.0}}$} & \multicolumn{1}{l}{69.4$_{\tiny\textcolor{green}{-7.2}}$} & \multicolumn{1}{l}{72.9$_{\tiny\textcolor{green}{-8.6}}$} \\
CoMVC \cite{DBLP:conf/cvpr/TrostenLJK21} & \multicolumn{1}{l}{86.9$_{\tiny\textcolor{red}{+6.0}}$} & \multicolumn{1}{l}{84.6$_{\tiny\textcolor{red}{+5.7}}$} & \multicolumn{1}{l}{98.5$_{\tiny\textcolor{red}{+17.6}}$} & \multicolumn{1}{l}{97.4$_{\tiny\textcolor{red}{+18.5}}$} & \multicolumn{1}{l}{98.1$_{\tiny\textcolor{red}{+21.5}}$} & \multicolumn{1}{l}{97.8$_{\tiny\textcolor{red}{+16.3}}$} & \multicolumn{1}{l}{90.6$_{\tiny\textcolor{red}{+14.0}}$} & \multicolumn{1}{l}{93.6$_{\tiny\textcolor{red}{+12.1}}$} \\
DIMVC \cite{xu2022deep} & \multicolumn{1}{l}{88.7$_{\tiny\textcolor{red}{+7.8}}$} & \multicolumn{1}{l}{93.7$_{\tiny\textcolor{red}{+14.8}}$} & \multicolumn{1}{l}{97.6$_{\tiny\textcolor{red}{+16.7}}$} & \multicolumn{1}{l}{96.0$_{\tiny\textcolor{red}{+17.1}}$} & \multicolumn{1}{l}{93.4$_{\tiny\textcolor{red}{+16.8}}$} & \multicolumn{1}{l}{93.5$_{\tiny\textcolor{red}{+12.0}}$} & \multicolumn{1}{l}{89.0$_{\tiny\textcolor{red}{+12.4}}$} & \multicolumn{1}{l}{91.7$_{\tiny\textcolor{red}{+10.2}}$} \\
DSMVC \cite{DBLP:conf/cvpr/Tang022} & \multicolumn{1}{l}{73.7$_{\tiny\textcolor{green}{-7.2}}$} & \multicolumn{1}{l}{72.2$_{\tiny\textcolor{green}{-6.7}}$} & \multicolumn{1}{l}{82.0$_{\tiny\textcolor{red}{+1.1}}$} & \multicolumn{1}{l}{81.4$_{\tiny\textcolor{red}{+2.5}}$} & \multicolumn{1}{l}{90.8$_{\tiny\textcolor{red}{+14.2}}$} & \multicolumn{1}{l}{96.5$_{\tiny\textcolor{red}{+15.0}}$} & \multicolumn{1}{l}{81.8$_{\tiny\textcolor{red}{+5.2}}$} & \multicolumn{1}{l}{84.1$_{\tiny\textcolor{red}{+2.6}}$} \\
DSIMVC \cite{DBLP:conf/icml/TangL22} & \multicolumn{1}{l}{90.4$_{\tiny\textcolor{red}{+9.5}}$} & \multicolumn{1}{l}{90.5$_{\tiny\textcolor{red}{+11.6}}$} & \multicolumn{1}{l}{99.0$_{\tiny\textcolor{red}{+18.1}}$} & \multicolumn{1}{l}{97.1$_{\tiny\textcolor{red}{+18.2}}$} & \multicolumn{1}{l}{99.7$_{\tiny\textcolor{red}{+23.1}}$} & \multicolumn{1}{l}{99.0$_{\tiny\textcolor{red}{+17.5}}$} & \multicolumn{1}{l}{98.8$_{\tiny\textcolor{red}{+22.2}}$} & \multicolumn{1}{l}{97.8$_{\tiny\textcolor{red}{+16.3}}$} \\
CPSPAN \cite{DBLP:conf/cvpr/JinWDLZ23} & \multicolumn{1}{l}{11.8$_{\tiny\textcolor{green}{-69.1}}$} & \multicolumn{1}{l}{0.3$_{\tiny\textcolor{green}{-78.6}}$} & \multicolumn{1}{l}{84.8$_{\tiny\textcolor{red}{+3.9}}$} & \multicolumn{1}{l}{82.1$_{\tiny\textcolor{red}{+3.2}}$} & \multicolumn{1}{l}{80.4$_{\tiny\textcolor{red}{+3.8}}$} & \multicolumn{1}{l}{85.1$_{\tiny\textcolor{red}{+3.6}}$} & \multicolumn{1}{l}{15.8$_{\tiny\textcolor{green}{-60.8}}$} & \multicolumn{1}{l}{3.3$_{\tiny\textcolor{green}{-78.2}}$} \\
SDMVC \cite{DBLP:journals/tkde/XuRTYPYPYH23} & \multicolumn{1}{l}{75.8$_{\tiny\textcolor{green}{-5.1}}$} & \multicolumn{1}{l}{72.2$_{\tiny\textcolor{green}{-6.7}}$} & \multicolumn{1}{l}{99.8$_{\tiny\textcolor{red}{+18.9}}$} & \multicolumn{1}{l}{99.5$_{\tiny\textcolor{red}{+20.6}}$} & \multicolumn{1}{l}{97.0$_{\tiny\textcolor{red}{+20.4}}$} & \multicolumn{1}{l}{95.6$_{\tiny\textcolor{red}{+14.1}}$} & \multicolumn{1}{l}{81.0$_{\tiny\textcolor{red}{+4.4}}$} & \multicolumn{1}{l}{89.2$_{\tiny\textcolor{red}{+7.7}}$} \\
MVCAN \cite{xu2024investigating} & \multicolumn{1}{l}{99.0$_{\tiny\textcolor{red}{+18.1}}$} & \multicolumn{1}{l}{98.4$_{\tiny\textcolor{red}{+19.5}}$} & \multicolumn{1}{l}{99.5$_{\tiny\textcolor{red}{+18.6}}$} & \multicolumn{1}{l}{98.8$_{\tiny\textcolor{red}{+19.9}}$} & \multicolumn{1}{l}{99.6$_{\tiny\textcolor{red}{+23.0}}$} & \multicolumn{1}{l}{99.1$_{\tiny\textcolor{red}{+17.6}}$} & \multicolumn{1}{l}{99.2$_{\tiny\textcolor{red}{+22.6}}$} & \multicolumn{1}{l}{98.8$_{\tiny\textcolor{red}{+17.3}}$} \\
CE-MVC [ours] & \multicolumn{1}{l}{99.7$_{\tiny\textcolor{red}{+18.8}}$} & \multicolumn{1}{l}{99.1$_{\tiny\textcolor{red}{+20.2}}$} & \multicolumn{1}{l}{99.6$_{\tiny\textcolor{red}{+18.7}}$} & \multicolumn{1}{l}{98.9$_{\tiny\textcolor{red}{+20.0}}$} & \multicolumn{1}{l}{99.9$_{\tiny\textcolor{red}{+23.3}}$} & \multicolumn{1}{l}{99.9$_{\tiny\textcolor{red}{+18.4}}$} & \multicolumn{1}{l}{99.8$_{\tiny\textcolor{red}{+23.2}}$} & \multicolumn{1}{l}{99.9$_{\tiny\textcolor{red}{+18.4}}$} \\
\hline
\end{tabular}
\label{1}
\end{table*}

\begin{table}[ht]
\centering
\caption{Comparison of different methods on real-world datasets.}
\label{tab:comparison}

\setlength{\tabcolsep}{2.0mm}
\begin{tabular}{lcccccccc}
\hline
\multirow{2}{*}{Method} & \multicolumn{2}{c}{RGB-D} & \multicolumn{2}{c}{Caltech} \\
\cmidrule(lr){2-3} \cmidrule(lr){4-5}
 & ACC & NMI & ACC & NMI \\
\hline
DEC-BestV \cite{DBLP:conf/icml/XieGF16} & \multicolumn{1}{l}{43.6} & \multicolumn{1}{l}{40.1} & \multicolumn{1}{l}{88.2} & \multicolumn{1}{l}{81.6} \\
DEC-WorstV \cite{DBLP:conf/icml/XieGF16} & \multicolumn{1}{l}{15.0$_{\tiny\textcolor{green}{-28.6}}$} & \multicolumn{1}{l}{5.1$_{\tiny\textcolor{green}{-35.0}}$} & \multicolumn{1}{l}{35.4$_{\tiny\textcolor{green}{-52.8}}$} & \multicolumn{1}{l}{19.6$_{\tiny\textcolor{green}{-62.0}}$} \\
DMJC \cite{DBLP:journals/tkde/XieLQLZMWT21} & \multicolumn{1}{l}{31.7$_{\tiny\textcolor{green}{-11.9}}$} & \multicolumn{1}{l}{28.5$_{\tiny\textcolor{green}{-11.6}}$} & \multicolumn{1}{l}{83.1$_{\tiny\textcolor{green}{-5.1}}$} & \multicolumn{1}{l}{80.3$_{\tiny\textcolor{green}{-1.3}}$} \\
DIMC-net \cite{DBLP:conf/mm/0001ZZWFXZ20} & \multicolumn{1}{l}{35.6$_{\tiny\textcolor{green}{-8.0}}$} & \multicolumn{1}{l}{32.4$_{\tiny\textcolor{green}{-7.7}}$} & \multicolumn{1}{l}{75.0$_{\tiny\textcolor{green}{-13.2}}$} & \multicolumn{1}{l}{68.5$_{\tiny\textcolor{green}{-13.1}}$} \\
GP-MVC \cite{DBLP:journals/tip/WangDTGF21} & \multicolumn{1}{l}{38.5$_{\tiny\textcolor{green}{-5.1}}$} & \multicolumn{1}{l}{32.6$_{\tiny\textcolor{green}{-7.5}}$} & \multicolumn{1}{l}{80.3$_{\tiny\textcolor{green}{-7.9}}$} & \multicolumn{1}{l}{77.6$_{\tiny\textcolor{green}{-4.0}}$} \\
CoMVC \cite{DBLP:conf/cvpr/TrostenLJK21} & \multicolumn{1}{l}{42.0$_{\tiny\textcolor{green}{-1.6}}$} & \multicolumn{1}{l}{41.3$_{\tiny\textcolor{red}{+1.2}}$} & \multicolumn{1}{l}{72.5$_{\tiny\textcolor{green}{-15.7}}$} & \multicolumn{1}{l}{68.8$_{\tiny\textcolor{green}{-12.8}}$} \\
DIMVC \cite{xu2022deep} & \multicolumn{1}{l}{46.9$_{\tiny\textcolor{red}{+3.3}}$} & \multicolumn{1}{l}{41.4$_{\tiny\textcolor{red}{+1.3}}$} & \multicolumn{1}{l}{87.2$_{\tiny\textcolor{green}{-1.0}}$} & \multicolumn{1}{l}{80.7$_{\tiny\textcolor{red}{-0.9}}$} \\
DSMVC \cite{DBLP:conf/cvpr/Tang022} & \multicolumn{1}{l}{43.3$_{\tiny\textcolor{green}{-0.3}}$} & \multicolumn{1}{l}{40.6$_{\tiny\textcolor{red}{+0.5}}$} & \multicolumn{1}{l}{90.5$_{\tiny\textcolor{red}{+2.3}}$} & \multicolumn{1}{l}{84.7$_{\tiny\textcolor{red}{+3.1}}$} \\
DSIMVC \cite{DBLP:conf/icml/TangL22} & \multicolumn{1}{l}{45.8$_{\tiny\textcolor{red}{+2.2}}$} & \multicolumn{1}{l}{41.0$_{\tiny\textcolor{red}{+0.9}}$} & \multicolumn{1}{l}{76.7$_{\tiny\textcolor{green}{-11.5}}$} & \multicolumn{1}{l}{67.5$_{\tiny\textcolor{green}{-14.1}}$} \\
CPSPAN \cite{DBLP:conf/cvpr/JinWDLZ23} & \multicolumn{1}{l}{42.4$_{\tiny\textcolor{green}{-1.2}}$} & \multicolumn{1}{l}{38.3$_{\tiny\textcolor{green}{-1.8}}$} & \multicolumn{1}{l}{84.8$_{\tiny\textcolor{green}{-3.4}}$} & \multicolumn{1}{l}{73.9$_{\tiny\textcolor{green}{-7.7}}$} \\
SDMVC \cite{DBLP:journals/tkde/XuRTYPYPYH23} & \multicolumn{1}{l}{44.1$_{\tiny\textcolor{red}{+0.5}}$} & \multicolumn{1}{l}{40.7$_{\tiny\textcolor{red}{+0.6}}$} & \multicolumn{1}{l}{88.3$_{\tiny\textcolor{green}{-2.9}}$} & \multicolumn{1}{l}{79.1$_{\tiny\textcolor{green}{-2.5}}$} \\
MVCAN \cite{xu2024investigating} & \multicolumn{1}{l}{48.0$_{\tiny\textcolor{red}{+4.4}}$} & \multicolumn{1}{l}{41.7$_{\tiny\textcolor{red}{+1.6}}$} & \multicolumn{1}{l}{93.6$_{\tiny\textcolor{red}{+5.4}}$} & \multicolumn{1}{l}{88.7$_{\tiny\textcolor{red}{+7.1}}$} \\
CE-MVC [ours] & \multicolumn{1}{l}{49.6$_{\tiny\textcolor{red}{+6.0}}$} & \multicolumn{1}{l}{42.6$_{\tiny\textcolor{red}{+2.5}}$} & \multicolumn{1}{l}{93.8$_{\tiny\textcolor{red}{+5.6}}$} & \multicolumn{1}{l}{89.1$_{\tiny\textcolor{red}{+7.5}}$} \\
\hline
\end{tabular}
\label{2}
\end{table}

\begin{figure*}[ht]
    \centering
    \begin{minipage}[b]{0.24\textwidth}
        \centering
        \captionsetup{labelformat=empty}
        \includegraphics[width=\textwidth]{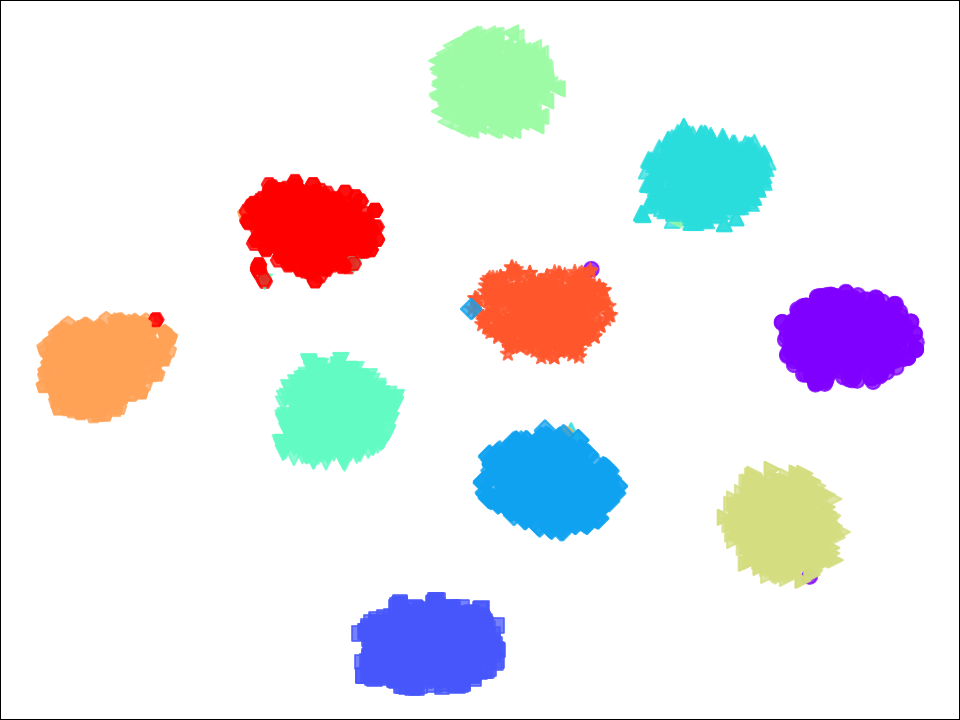}
        \subcaption{DIGIT}
    \end{minipage}
    \hfill
    \begin{minipage}[b]{0.24\textwidth}
        \centering
        \captionsetup{labelformat=empty}
        \includegraphics[width=\textwidth]{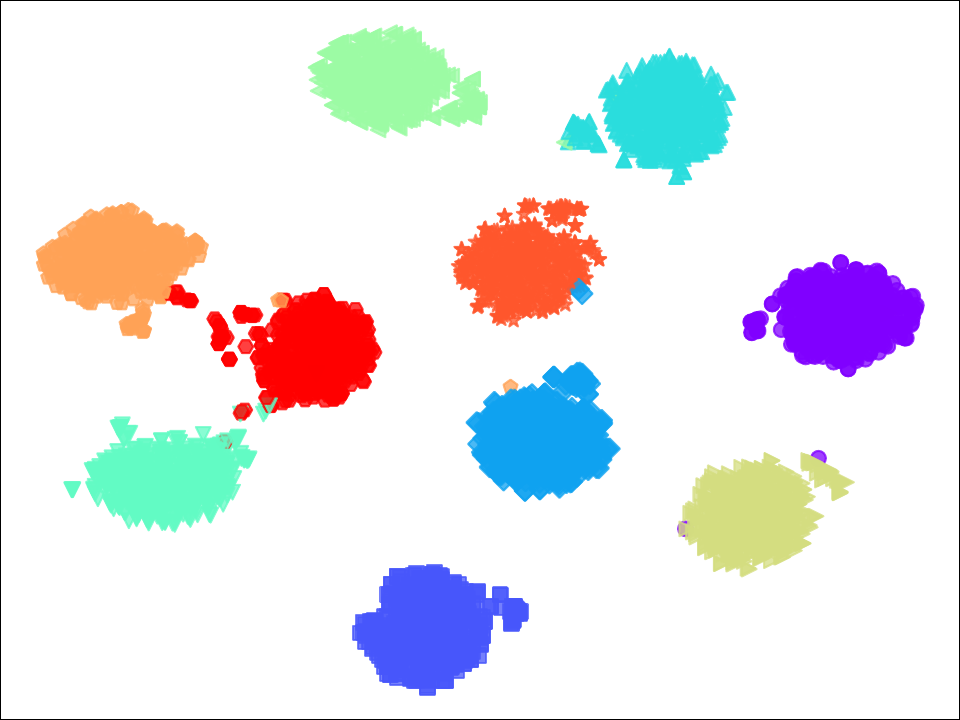}
        \subcaption{NoisyDIGIT}
    \end{minipage}
    \hfill
    \begin{minipage}[b]{0.24\textwidth}
        \centering
        \captionsetup{labelformat=empty}
        \includegraphics[width=\textwidth]{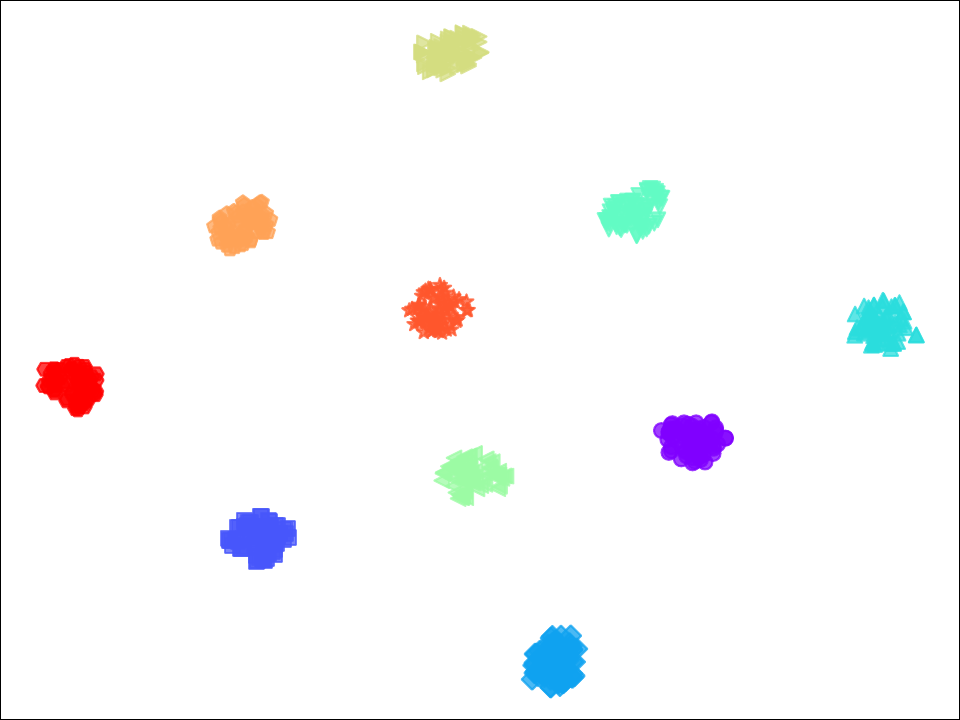}
        \subcaption{COIL}
    \end{minipage}
    \hfill
    \begin{minipage}[b]{0.24\textwidth}
        \centering
        \captionsetup{labelformat=empty}
        \includegraphics[width=\textwidth]{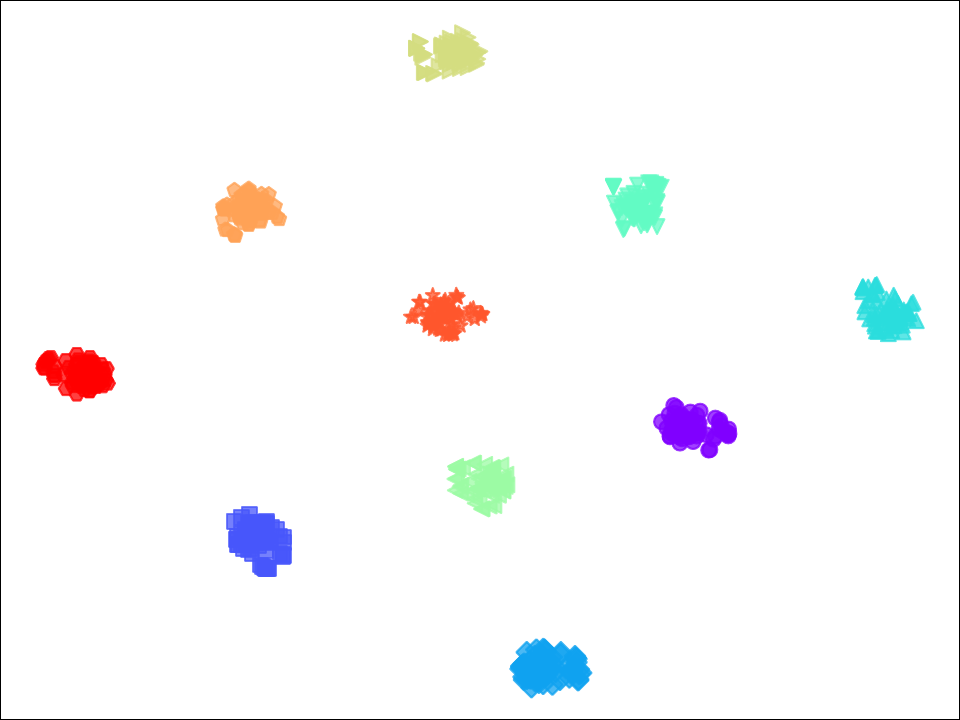}
        \subcaption{NoisyCOIL}
    \end{minipage}
    \caption{The figure illustrates the visualization for CE-MVC framework's clustering results on DIGIT and COIL datasets as well as their noisy-contaminated versions, demonstrating the robust clustering performance of the proposed model.}
    \label{f}
\end{figure*}

\section{Experiments}
\subsection{Experimental setup}
\subsubsection{Datasets}
To evaluate the performance of the proposed framework, we conduct experiments on two normal multi-view clustering datasets (including DIGIT and COIL) which are easy for clustering, as well as their noise-contaminated versions. The noise-contaminated datasets are constructed by constructing an additional view with randomly sampled noise for each dataset, effectively testing the robustness of our model under extreme circumstances. In addition, several real-world multi-view datasets that are challenging for clustering are utilized to further verify the performance, such as RGB-D and Caltech, which are hard for clustering.
\subsubsection{Compared methods}
We compare our proposed CE-MVC with ten state-of-the-art self-supervised clustering algorithms. Specifically, DEC \cite{DBLP:conf/icml/XieGF16} is a well-established deep SVC method that we use as a baseline to assess the impact of NVD on MVC methods. Among the compared methods, DMJC \cite{DBLP:journals/tkde/XieLQLZMWT21}, DIMC-net \cite{DBLP:conf/mm/0001ZZWFXZ20}, GP-MVC \cite{DBLP:journals/tip/WangDTGF21}, DIMVC \cite{xu2022deep}, and SDMVC \cite{DBLP:journals/tkde/XuRTYPYPYH23} are deep MVC approaches that extend DEC, often employing consistent soft labels to ensure clustering consistency. Additionally, DMJC \cite{DBLP:journals/tkde/XieLQLZMWT21}, DIMC-net \cite{DBLP:conf/mm/0001ZZWFXZ20}, GP-MVC \cite{DBLP:journals/tip/WangDTGF21}, and DSMVC \cite{DBLP:conf/cvpr/Tang022} incorporate weighting strategies to derive fused representations. On the other hand, CoMVC \cite{DBLP:conf/cvpr/TrostenLJK21}, DSIMVC \cite{DBLP:conf/icml/TangL22}, and CPSPAN \cite{DBLP:conf/cvpr/JinWDLZ23} are deep MVC methods based on contrastive learning which is capable of learning shared representations across multiple views.

\subsection{Comparison Results and Analysis}
Tables \ref{1} and \ref{2} provide a comprehensive comparison of clustering performance across various datasets, evaluated using clustering accuracy (ACC) and normalized mutual information (NMI). DEC-BestV and DEC-WorstV represent the results of the SVC method DEC on the best and worst views, respectively. The significant performance drop observed in DEC-WorstV across all datasets, such as the 68.5\% decrease in ACC and 78.5 in NMI on the NoisyDIGIT dataset, highlights the variability in view quality and the challenges posed by noisy views in multi-view clustering. 
% Most MVC methods show performance improvements over DEC-BestV on normal datasets, as evidenced by DMJC's slight degradation in ACC (+0.2\%) on NoisyDIGIT and CoMVC's substantial increases in ACC (+6.0\%) and NMI (+5.7) on the same dataset. 
However, on datasets like COIL and NoisyCOIL, some MVC methods, including GP-MVC, exhibit performance degradation compared to DEC-BestV, with ACC dropping by 31.8\% on NoisyDIGIT and NMI by 15.4. Despite these challenges, CE-MVC consistently outperforms DEC-BestV and most other MVC methods, demonstrating significant improvements across all datasets, such as a 23.3\% increase in ACC and an 18.4 boost in NMI on the COIL dataset. In simulated noisy environments, CE-MVC further proves its robustness by surpassing the best comparison methods with substantial gains, particularly on NoisyDIGIT, where it achieves an 18.8\% improvement in ACC and 20.2 in NMI. Additionally, CE-MVC consistently outperforms all other methods across two real-world datasets, achieving the highest ACC and NMI scores. Particularly on the Caltech dataset, CE-MVC improves ACC by 5.6\% and NMI by 7.5. In contrast, traditional single-view methods like DEC-WorstV show significantly poor performance in noisy view scenarios, with notable drops in both ACC and NMI. This consistent performance superiority indicates CE-MVC's ability to effectively handle noisy views and extract useful, complementary information, making it highly effective in both simulated and real-world multi-view clustering scenarios.
% Tables \ref{1} and \ref{2} compare clustering performance across various datasets, using accuracy (ACC) and normalized mutual information (NMI) as metrics. DEC-BestV and DEC-WorstV represent the results of the SVC method DEC on the best and worst views, respectively. DEC-WorstV shows significant performance drops across all datasets, such as a 68.5\% decrease in ACC and 78.5 in NMI on NoisyDIGIT, highlighting the variability in view quality and the challenges noisy views pose in multi-view clustering. Most MVC methods improve upon DEC-BestV on normal datasets, with DMJC showing a slight ACC gain (+0.2\%) on NoisyDIGIT, and CoMVC achieving substantial improvements (+6.0\% ACC, +5.7 NMI). However, on datasets like COIL and NoisyCOIL, some methods, such as GP-MVC, show performance degradation compared to DEC-BestV, with ACC dropping by 31.8\% and NMI by 15.4 on NoisyDIGIT. Despite these challenges, CE-MVC consistently outperforms DEC-BestV and other MVC methods, showing notable gains across all datasets, including a 23.3\% increase in ACC and 18.4 in NMI on COIL. In noisy environments, such as NoisyDIGIT, CE-MVC demonstrates strong robustness, achieving an 18.8\% improvement in ACC and 20.2 in NMI. Additionally, on real-world datasets like RGB-D and Caltech, CE-MVC delivers the highest ACC and NMI scores, with a 5.6\% improvement in ACC and 7.5 in NMI on Caltech. These results highlight CE-MVC’s consistent ability to handle noisy views and extract complementary information, making it highly effective in both simulated and real-world multi-view clustering scenarios.

To further validate the effectiveness of the CE-MVC framework, we visualize the clustering results on the DIGIT and COIL datasets, as well as their corresponding noise-contaminated versions, as shown in Figure \ref{f}. It can be observed that despite the noise introducing some disturbance to the clustering boundaries, the model is still able to achieve distinguishable clustering boundaries, suggesting the robustness of the CE-MVC framework under challenging conditions.
\subsection{Ablation study}
\begin{table}[htbp]
    \centering
    \renewcommand\arraystretch{1}
    \setlength{\tabcolsep}{2.5mm}
    \caption{Ablation study on distinct weighting strategies}
    \begin{tabular}{ccc|cc|cc}
    \toprule
    \multicolumn{3}{c|}{Weighting} & \multicolumn{2}{c|}{RGB-D} & \multicolumn{2}{c}{Caltech}     \\
    \cmidrule(lr){1-3} \cmidrule(lr){4-5} \cmidrule(lr){6-7}
    NMI & ENMI& CE & ACC & NMI & ACC & NMI \\
    \midrule
    \checkmark &  &  & 44.5 & 40.3 & 91.1 & 83.7 \\
     & \checkmark &  & 47.6 & 40.8 &  92.5 & 86.6\\
     % \checkmark &  & \checkmark & 49.6 & 42.6 & 93.8 & 89.1\\
     & \checkmark & \checkmark & 49.6 & 42.6 & 93.8 & 89.1\\
    \hline
\end{tabular}
\label{memory}
\end{table}
Table \ref{memory} presents the results of an ablation study conducted to assess the impact of different weighting strategies on the performance of the CE-MVC framework, on RGB-D and Caltech datasets. The table evaluates three distinct strategies: NMI, Exponential NMI (ENMI), and Conditional Entropy (CE). It can be observed that when both ENMI and CE are employed together as weighting strategies, the results show further improvement, particularly evident in the RGB-D dataset where ACC and NMI increase to 49.6\% and 42.6\%, respectively, which confirms that simultaneously considering NMI and conditional entropy metrics allows for better utilization of valuable consistency and complementarity information from different views for clustering.

\section{Conclusion}
This study proposes CE-MVC, a robust multi-view clustering framework that optimizes clustering performance by addressing view consistency, complementarity, and noise. Using parameter-decoupled model and adaptive weighting based on conditional entropy and NMI, CE-MVC outperforms existing methods, particularly in noisy settings, making it an effective solution for complex clustering tasks.
% This study introduces the CE-MVC framework, a robust multi-view clustering approach that effectively addresses the challenges of view consistency, complementarity, and noise interference. By integrating adaptive weighting strategies based on conditional entropy and NMI, the framework enhances clustering accuracy across diverse datasets. Experimental results demonstrate CE-MVC's superior performance and resilience, particularly in noisy environments, establishing it as a highly effective solution for complex multi-view clustering tasks.
% \section*{Acknowledgment}

% \section*{References}
\bibliographystyle{unsrt}
\bibliography{IEEEfull.bib}
\end{document}